\pdfoutput=1
\documentclass[10pt]{article}

\usepackage[letterpaper,margin=1in]{geometry}
\usepackage{microtype}
\usepackage{times}
\usepackage[T1]{fontenc}

\usepackage[numbers,sort&compress]{natbib}
\usepackage[hidelinks,colorlinks=false,hyperfootnotes=false]{hyperref}
\usepackage{xurl}

\usepackage{booktabs}
\usepackage{array}
\usepackage{tabularx}
\usepackage{multirow}
\usepackage{graphicx}
\usepackage{caption}
\usepackage{amsmath}
\usepackage{amssymb}
\usepackage{mathtools}
\usepackage{bm}

\usepackage{xcolor}
\usepackage{tikz}
\usepackage{pgfplots}
\usepgfplotslibrary{fillbetween}
\pgfplotsset{compat=1.18}
\usetikzlibrary{patterns,positioning,arrows.meta,calc,shapes.geometric,fit,backgrounds}

\usepackage{pifont}
\usepackage{enumitem}
\setlist{nosep,leftmargin=1.4em}

\definecolor{cLLaVA}{HTML}{2E4F7A}      % deep navy (primary)
\definecolor{cPaliGemma}{HTML}{D97757}   % Anthropic coral (accent/secondary)
\definecolor{cQwen}{HTML}{8C5E3C}        % warm sepia/umber (tertiary)
\definecolor{cAccent}{HTML}{D97757}      % coral accent
\definecolor{cMuted}{HTML}{7A7268}       % warm grey
\definecolor{cBg}{HTML}{F5F2EB}          % cream panel background
\definecolor{cBgDeep}{HTML}{EDE8DC}      % deeper cream for emphasis
\definecolor{cInk}{HTML}{3B342C}         % dark sepia ink
\definecolor{cGrid}{HTML}{FFFFFF}        % white gridlines on cream

\newcolumntype{L}[1]{>{\raggedright\arraybackslash}p{#1}}
\newcolumntype{C}[1]{>{\centering\arraybackslash}p{#1}}
\newcolumntype{Y}{>{\raggedright\arraybackslash}X}

\newcommand{\method}{\textsc{Vrp}\xspace}
\usepackage{xspace}
\usepackage{float}
\usepackage{placeins}
\newcommand{\eg}{e.g.,\xspace}
\newcommand{\ie}{i.e.,\xspace}

\newcommand{\Hs}{H_{\mathrm{s}}}
\newcommand{\Ck}{C_{k}}
\newcommand{\Ktot}{K_{\mathrm{tot}}}
\newcommand{\AUROC}{\mathrm{AUROC}}
\newcommand{\Rpb}{R_{\mathrm{pb}}}
\newcommand{\eqlabel}[1]{\label{eq:#1}}

\title{\vspace{-1.5em}\textbf{Where Reliability Lives in Vision--Language Models:\\
A Mechanistic Study of Attention, Hidden States, and Causal Circuits}\vspace{-0.3em}}

\author{
\textbf{Logan Mann}$^{1,\ast}$ \quad
\textbf{Ajit Saravanan}$^{1}$ \quad
\textbf{Ishan Dave}$^{2}$ \quad
\textbf{Shikhar Shiromani}$^{3}$ \\[2pt]
\textbf{Saadullah Ismail}$^{4}$ \quad
\textbf{Yi Xia}$^{4}$ \quad
\textbf{Emily Huang}$^{5}$ \\[4pt]
\small $^{1}$UC Santa Barbara \quad
$^{2}$UC Berkeley \quad
$^{3}$NVIDIA \quad
$^{4}$Algoverse AI Research \quad
$^{5}$Brown University \\[2pt]
\small $^{\ast}$Correspondence: \texttt{loganmann@ucsb.edu}
}
\date{}

\begin{document}
\maketitle
\vspace{-1.5em}
\begingroup
\renewcommand{\thefootnote}{}
\footnotetext{Accepted at the \emph{ICLR 2026 Workshop on Multimodal Reasoning}.}
\addtocounter{footnote}{-1}
\endgroup

\begin{abstract}
A pervasive intuition holds that vision--language models (VLMs) are most trustworthy when their attention maps look sharp: concentrated attention on the queried region should imply a confident, calibrated answer. We test this \emph{Attention--Confidence Assumption} directly. We instrument three open-weight VLM families (LLaVA-1.5, PaliGemma, Qwen2-VL; 3--7B parameters) with a unified mechanistic pipeline---the \emph{VLM Reliability Probe} (\method)---that compares attention structure, generation dynamics, and hidden-state geometry against a single correctness label. Three results emerge. \textbf{(i)} Attention structure is a near-zero predictor of correctness ($\Rpb(\Ck,y){=}0.001$, $95\%$\,CI~$[-0.034,0.036]$; $\Rpb(\Hs,y){=}{-}0.012$, $[-0.047,0.024]$ on a pooled $n{=}3{,}090$ split), even though attention remains \emph{causally} necessary for feature extraction (top-30\% patch masking drops accuracy by $8.2$--$11.3$\,pp, $p{<}0.001$). \textbf{(ii)} Reliability becomes legible later in the computation: a single hidden-state linear probe reaches $\AUROC{>}0.95$ on POPE for two of three families, and self-consistency at $K{=}10$ is the strongest behavioral predictor we measure at $10\times$ inference cost ($\Rpb{=}0.43$). \textbf{(iii)} Causal neuron-level ablations expose a sharp architectural split with direct monitor-design implications: late-fusion LLaVA concentrates reliability in a fragile late bottleneck ($-8.3$\,pp object-identification accuracy after top-5 probe-neuron ablation), whereas early-fusion PaliGemma and Qwen2-VL distribute it widely and absorb destruction of $\sim$50\% of their peak-layer hidden dimension with $\le 1$\,pp degradation. The takeaway is narrow but consequential: in 3--7B VLMs, reliability is read more reliably off hidden-state geometry, layer-wise margin formation, and sparse late-layer circuits than off attention-map sharpness.
\end{abstract}

\section{Introduction}
\label{sec:intro}

Vision--language models can answer richly compositional questions about images,
yet routinely produce \emph{fluent} mistakes: confident, well-formed answers
that are not supported by the pixels they purport to describe
\citep{liu2023visual,beyer2024paligemma,wang2024qwen2}. For deployment in
settings where errors carry cost (scientific image analysis, medical triage,
robotic perception), we need reliability signals that are simultaneously
\emph{predictive of correctness} and \emph{mechanistically interpretable}.
This raises a sharp interpretability question: where, inside a VLM, is the
information that distinguishes a correct answer from an incorrect one?

A natural and visually intuitive hypothesis is that reliability lives in
attention. Cross-attention maps are easy to extract, easy to visualize, and are
frequently treated as a window onto what the model ``used'' to produce its
answer \citep{jain2019attention,wiegreffe2019attention}. We refer to the
operationalization of this intuition as the \emph{Attention--Confidence
Assumption}: \emph{if a VLM concentrates its visual attention on the relevant
region, the resulting answer should be more trustworthy; diffuse attention
should signal lower reliability}. The Attention--Confidence Assumption is
strictly stronger than the (well-supported) claim that attention is causally
involved in computation. It additionally requires that the \emph{structure} of
attention (its sharpness, fragmentation, or entropy) be calibrated to the
model's probability of being right.

We test this assumption head-on. We introduce the \emph{VLM Reliability Probe}
(\method), a unified mechanistic pipeline that instruments three open VLM
families (LLaVA-1.5-7B, PaliGemma-3B, Qwen2-VL-7B) and compares attention
structure against generation dynamics and hidden-state readouts on the same
inputs and the same correctness labels. \method\ extracts cross-attention
tensors, hidden states, and per-token confidences via forward hooks; reduces
attention to per-layer spatial vectors and structural summaries (entropy $\Hs$,
secondary-component count $\Ck$); applies the logit lens
\citep{nostalgebraist2020logit} to track when the correct token first separates
from competitors in the residual stream; trains $L_1$-regularized linear probes
to localize sparse reliability circuits; and validates findings with targeted
neuron ablation and patch masking.

\paragraph{Findings.}
Three results emerge across families. (i)~Attention \emph{structure} is a
near-zero predictor of correctness, even though attention remains causally
necessary for feature extraction; a supervised non-linear ensemble over 32
attention layers tops out at $\AUROC{=}0.725$. (ii)~Reliability becomes
legible only later: the logit-lens truth margin peaks deep in the stack and
is dominated by MLP residual contributions ($\sim 70$--$82$\%), and single
hidden-state probes reach $\AUROC{>}0.95$ on POPE for LLaVA and Qwen2-VL.
(iii)~Architectures organize this signal differently---LLaVA concentrates it
in a fragile late bottleneck, whereas PaliGemma and Qwen2-VL distribute it
across a wide manifold robust to massive ablation.

\paragraph{Contributions.}
We (i) pose and falsify the Attention--Confidence Assumption under a uniform
protocol across three VLM families and four benchmarks; (ii) map \emph{when
and where} reliability becomes linearly decodable using logit-lens
trajectories, $L_1$-regularized neuron probes, and residual-update analysis;
(iii) provide causal evidence---negative (top-$k$ and random ablation, MLP
bypass) and positive (top-30\% patch masking)---that the located circuit is
not merely correlational, and document a sharp robustness asymmetry across
families; and (iv) extend a probing literature
\citep{burns2023ccs,marks2024geometry,geva2021kv} so far applied mostly to
text-only models, arguing that VLM monitor design should prefer hidden-state
and consistency-based signals over attention-map heuristics.

Code, prompts, split definitions, and probe-training pipelines are released at
\href{https://github.com/itsloganmann/VLM-Reliability-Probe}{\nolinkurl{github.com/itsloganmann/VLM-Reliability-Probe}}.

\section{Related Work}
\label{sec:related}

\paragraph{Vision--language models and hallucination benchmarks.}
Large VLMs build on contrastive and encoder--decoder vision--language
pretraining combined with strong language backbones, enabling instruction
following and open-ended multimodal generation
\citep{radford2021clip,li2022blip,alayrac2022flamingo,liu2023visual,dai2023instructblip,beyer2024paligemma,wang2024qwen2}.
Their fluency makes reliability difficult to judge: models produce confident
answers that are weakly grounded in the image. This concern has motivated
benchmark-driven work on object hallucination and multimodal evaluation,
including POPE, LLaVA-Bench, MME, SEED-Bench, MM-Vet, and the CHAIR family
\citep{li2023pope,zhou2023llavabench,fu2023mme,li2023seedbench,yu2023mmvet,rohrbach2018object}.
These benchmarks establish \emph{where} models fail; they do not, by
themselves, locate \emph{where} the failure-relevant computation lives.

\paragraph{Attention as explanation.}
Whether attention is a faithful explanation of model behavior has been debated
in NLP \citep{jain2019attention,wiegreffe2019attention,serrano2019attention}.
For VLMs, recent evidence shows that correct localization and correct
answering can come apart: models often attend to the right region while
reasoning incorrectly about it \citep{liu2025seeing}. Saliency- and
attribution-based interpretability \citep{chefer2021genattr} provides finer
spatial maps, but the question of whether \emph{any} spatial summary of
attention predicts correctness has not been answered cleanly across families.
We target precisely that question.

\paragraph{Mechanistic interpretability and probing for truthfulness.}
A growing literature reads model state for evidence of correctness or
truthfulness. \citet{burns2023ccs} discover linear directions associated with
truthful belief in language models without supervision; \citet{marks2024geometry}
show that truthful and false statements separate along a low-dimensional
geometry in the residual stream; and \citet{geva2021kv,geva2022promote}
characterize the role of MLP layers as key--value memories that promote tokens
in the vocabulary space. The logit lens \citep{nostalgebraist2020logit} and
tuned lens variants \citep{belrose2023tunedlens} provide layer-wise readouts
of the residual stream. To date, these tools have been applied mostly to
text-only models. \citet{long2025lvlm} introduce a hidden-state perspective
on VLMs via the Visual Integration Point. Our work combines these
perspectives in an explicitly mechanistic pipeline that compares attention
structure, layer-wise hidden-state readouts, sparse unit-level probes, and
causal interventions within a single cross-family analysis of VLM reliability.

\paragraph{Behavioral reliability.}
Self-consistency \citep{wang2023selfconsistency} aggregates agreement across
sampled reasoning paths; semantic-entropy \citep{kuhn2023semantic} and
p(True) self-evaluation \citep{kadavath2022pmt} extend this to free-form
output. We include self-consistency as a strong behavioral baseline and
compare it directly against single-pass internal readouts.

\section{The VLM Reliability Probe}
\label{sec:method}

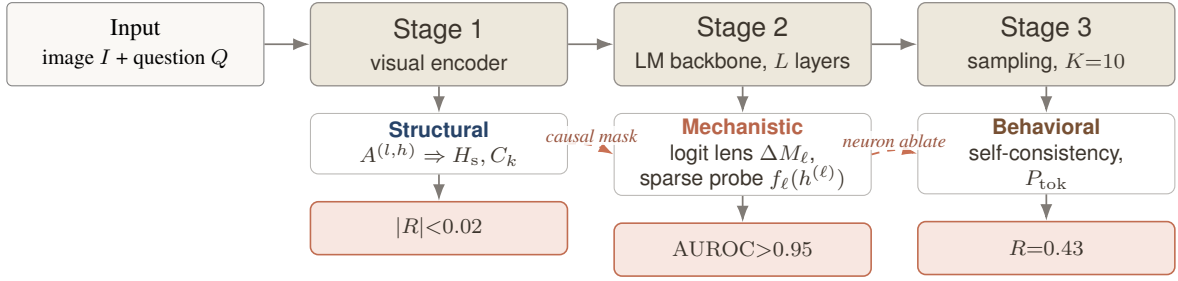
\begin{figure*}[t]
\centering
\begin{tikzpicture}[
  font=\small,
  node distance=3.5mm and 6mm,
  >=Latex,
  every node/.style={align=center},
  input/.style={rounded corners=2pt, draw=black!50, fill=cBg!50, minimum width=34mm, minimum height=11mm, inner sep=3pt},
  stage/.style={rounded corners=3pt, draw=cInk!55, line width=0.5pt, fill=cBgDeep, minimum width=34mm, minimum height=11mm, inner sep=3pt, text=cInk, font=\sffamily},
  metric/.style={rounded corners=3pt, draw=cInk!35, line width=0.4pt, fill=white, minimum width=34mm, minimum height=8mm, inner sep=2pt, font=\footnotesize\sffamily, text=cInk},
  result/.style={rounded corners=3pt, draw=cAccent!85!cInk, line width=0.6pt, fill=cAccent!18, minimum width=34mm, minimum height=7mm, inner sep=2pt, font=\footnotesize\sffamily\itshape, text=cInk},
  arr/.style={-{Latex[length=2.2mm,width=1.6mm]}, line width=0.65pt, cInk!65},
  cause/.style={-{Latex[length=2.0mm,width=1.4mm]}, line width=0.7pt, cAccent!90!cInk, dashed}
]
\node[input] (input) {Input\\\footnotesize image $I$ + question $Q$};
\node[stage, right=of input] (vis) {Stage 1\\\footnotesize visual encoder};
\node[stage, right=of vis] (lm) {Stage 2\\\footnotesize LM backbone, $L$ layers};
\node[stage, right=of lm] (samp) {Stage 3\\\footnotesize sampling, $K{=}10$};
\node[metric, below=of vis] (attn) {{\color{cLLaVA!85!black}\textbf{Structural}}\\$A^{(l,h)}\Rightarrow \Hs, \Ck$};
\node[metric, below=of lm] (mech) {{\color{cPaliGemma!85!black}\textbf{Mechanistic}}\\logit lens $\Delta M_{\ell}$,\\sparse probe $f_{\ell}(h^{(\ell)})$};
\node[metric, below=of samp] (beh) {{\color{cQwen!85!black}\textbf{Behavioral}}\\self-consistency,\\$P_{\mathrm{tok}}$};
\node[result, below=of attn] (rattn) {$|R|{<}0.02$};
\node[result, below=of mech] (rmech) {$\mathrm{AUROC}{>}0.95$};
\node[result, below=of beh] (rbeh) {$R{=}0.43$};
\draw[arr] (input) -- (vis);
\draw[arr] (vis) -- (lm);
\draw[arr] (lm) -- (samp);
\draw[arr] (vis) -- (attn);
\draw[arr] (lm) -- (mech);
\draw[arr] (samp) -- (beh);
\draw[arr] (attn) -- (rattn);
\draw[arr] (mech) -- (rmech);
\draw[arr] (beh) -- (rbeh);
% causal intervention edges (dashed orange) - labels above the arrows
\draw[cause] (attn.east) to[bend left=15] node[midway, above, font=\scriptsize\itshape, text=cAccent!70!black, fill=white, fill opacity=0.85, text opacity=1, inner sep=1pt] {causal mask} (mech.west);
\draw[cause] (mech.east) to[bend left=15] node[midway, above, font=\scriptsize\itshape, text=cAccent!70!black, fill=white, fill opacity=0.85, text opacity=1, inner sep=1pt] {neuron ablate} (beh.west);
\end{tikzpicture}
\caption{\textbf{The VLM Reliability Probe (\method).} A unified pipeline that extracts three classes of evidence on a common footing. Stage~1 reduces cross-attention to per-layer spatial vectors and structural summaries ($\Hs, \Ck$). Stage~2 reads the residual stream via the logit lens and $L_1$-sparse probes. Stage~3 samples $K{=}10$ outputs to compute self-consistency. Dashed orange edges denote causal interventions: top-30\% patch masking on attention and top-$k$ neuron ablation on the residual stream. Headline numbers below each metric family preview the central finding of \S5.}
\label{fig:vrp}
\end{figure*}

We instrument each model with forward hooks that record (i) cross-attention
tensors $A^{(l,h)} \in \mathbb{R}^{T \times S}$ at every decoder layer $l$ and
head $h$ (where $T$ is the number of generated answer tokens and $S$ is the
number of image patches), (ii) residual hidden states $h^{(\ell)} \in
\mathbb{R}^{d}$ at every layer, and (iii) per-token output probabilities. From
these signals we derive three families of metrics; see Figure~\ref{fig:vrp}.
The pipeline is designed to disentangle two competing hypotheses:
\begin{description}
\item[H1: Structural Hypothesis.] Reliability is grounded in the spatial
coherence of the visual encoder's attention, namely \emph{how the model looks}.
\item[H2: Mechanistic--Consistency Hypothesis.] Reliability emerges from
generation dynamics and the geometry of late-layer hidden states, namely \emph{what
the model is converging toward}.
\end{description}

\subsection{Stage 1: Structural Metrics from Attention}

For each layer $l$, we average $A^{(l,h)}$ over heads and over answer-token
positions to obtain a single spatial vector $m^{(l)} \in \mathbb{R}^{S}$ over
image patches, then normalize to a probability distribution $\tilde m^{(l)}$.
We summarize this distribution with two structural quantities:

\begin{align}
\Hs^{(l)} &= -\sum_{s=1}^{S} \tilde m^{(l)}_s \log \tilde m^{(l)}_s
  && \text{(spatial entropy)} \eqlabel{entropy} \\
\Ck^{(l)} &= \Ktot^{(l)} - 1
  && \text{(secondary-component count).} \eqlabel{clusters}
\end{align}

To compute $\Ktot^{(l)}$, we threshold $\tilde m^{(l)}$ at the top $30\%$ of attention mass, binarize on the patch grid, and count connected components under 4-neighbor adjacency, mirroring the saliency-thresholding convention used in attention-based interpretability \citep{chefer2021genattr}. $\Ktot^{(l)} = 1$ corresponds
to a single contiguous focus, hence $\Ck^{(l)} = 0$. Throughout the paper we
report $\Ck$ rather than $\Ktot$ unless explicitly noted, so that ``zero''
corresponds to the maximally focused case. We also track layer-wise
attention-evolution deltas $\Delta\Hs^{(l)} = \Hs^{(l)} - \Hs^{(l-1)}$ to
characterize how attention sharpens or diffuses through the stack. As a
robustness check, we re-run all attention analyses with a DBSCAN variant
($\varepsilon{=}1.5$, $\mathrm{min\_samples}{=}3$); results agree to within
$\pm0.01$ in $\Rpb$.

\subsection{Stage 2: Mechanistic Readouts via the Logit Lens and Probes}
\label{sec:method-mech}

Let $W_U \in \mathbb{R}^{|V| \times d}$ denote the unembedding matrix and let
$z_\ell = W_U \, \mathrm{LN}(h^{(\ell)}) \in \mathbb{R}^{|V|}$ be the layer-$\ell$
logit-lens projection \citep{nostalgebraist2020logit}, where $\mathrm{LN}$ is
the model's final-layer norm applied to the residual stream. We define the
\emph{truth margin}
\begin{equation}
\Delta M_{\ell} = z_{\ell}(y^{\star}) - \max_{y \neq y^{\star}} z_{\ell}(y),
\eqlabel{margin}
\end{equation}
where $y^{\star}$ is the reference answer token under our evaluation protocol
(\S\ref{sec:protocol}). For closed-form benchmarks (POPE, yes/no) $y^{\star}$
is unambiguous; for open-ended benchmarks we follow the protocol in
\S\ref{sec:protocol} and use the first content token of the canonicalized
ground-truth answer string, mirroring the convention adopted in recent
logit-lens analyses of multimodal models \citep{long2025lvlm}.

At every layer we additionally train a learned probe $f_{\ell}: \mathbb{R}^{d}
\to [0,1]$ predicting binary correctness from $h^{(\ell)}$ alone. We report
two variants: (a) a logistic probe with $L_2$ regularization (dense), and (b)
a logistic probe with $L_1$ regularization at $\lambda{=}0.1$ (sparse). The
sparse probe selects compact units that we use for the neuron-level and causal
ablation analyses in \S\ref{sec:circuits}. To attribute the layerwise growth
of $\Delta M_\ell$, we decompose the residual update at layer $\ell$ into its
MLP and attention contributions and report their relative magnitudes,
following \citet{geva2022promote}.

\subsection{Stage 3: Behavioral Metrics from Generation Dynamics}

For each example we draw $K{=}10$ samples $\{y_1, \dots, y_K\}$ under nucleus
sampling ($p{=}0.9$, $T{=}0.7$). We compute self-consistency as the support of
the majority answer:
\begin{equation}
\mathrm{SC} = \max_{a} \frac{1}{K} \sum_{k=1}^{K} \mathbf{1}[\,\Phi(y_k) = a\,],
\eqlabel{sc}
\end{equation}
where $\Phi$ is a canonicalization function that lower-cases, strips
punctuation, and applies benchmark-specific normalization (\eg yes/no
collapsing on POPE, integer extraction on counting). We additionally record the
single-pass token confidence $P_{\mathrm{tok}}$ assigned to the emitted answer
token, and, for free-form benchmarks, the geometric mean of token
probabilities up to the first newline. All structural, mechanistic, and
behavioral signals are evaluated against the same binary correctness labels
using $\Rpb$ and $\AUROC$.

\section{Experimental Protocol}
\label{sec:protocol}

\begin{table}[ht]
\centering
\small
\caption{\textbf{Models evaluated.} Three open-weight VLMs spanning late-fusion (LLaVA), early-fusion (PaliGemma), and dynamic-resolution early-fusion (Qwen2-VL) designs.}
\label{tab:models}
\begin{tabular}{@{}lllll@{}}
\toprule
\textbf{Model} & \textbf{Decoder} & \textbf{Vision encoder} & \textbf{Connector} & \textbf{Fusion} \\
\midrule
LLaVA-1.5-7B & Vicuna (32L) & CLIP ViT-L/14 (frozen) & 2-layer MLP & late \\
PaliGemma-3B & Gemma (18L) & SigLIP & linear projector & early \\
Qwen2-VL-7B-Instruct & Qwen2 (28L, GQA) & native multimodal tokens & --- & early, dynamic res. \\
\bottomrule
\end{tabular}
\end{table}

Table~\ref{tab:models} summarizes the three open-weight VLMs we evaluate \citep{liu2023visual,beyer2024paligemma,wang2024qwen2}, spanning late-fusion, early-fusion, and dynamic-resolution early-fusion designs. All experiments use HuggingFace implementations on NVIDIA A100-80GB GPUs.

\paragraph{Benchmarks.} We evaluate on:
(i)~\textbf{POPE}-Adversarial \citep{li2023pope}, $n{=}1{,}000$ binary
yes/no object-existence queries that stress object hallucination;
(ii)~\textbf{LLaVA-Bench} \citep{zhou2023llavabench}, $n{=}90$ open-ended
reasoning prompts;
(iii)~a \textbf{custom counting + spatial} suite of $n{=}2{,}000$ items
($1{,}000$ counting, $1{,}000$ spatial relations) constructed from COCO-style
images with manually verified integer / relation labels;
(iv)~\textbf{VQAv2}-val \citep{goyal2017vqav2} for general scene
understanding, and (v)~\textbf{TextVQA} \citep{singh2019textvqa} for
OCR-heavy questions. We report $\Rpb$ with binary correctness for primary
claims and $\AUROC$ for reliability prediction. Sample accounting and 95\%
bootstrap confidence intervals (10{,}000 resamples) for all headline
numbers are summarized in Table~\ref{tab:samples}.

\paragraph{Reference-token protocol.}
For closed-form benchmarks, $y^{\star}$ in Eq.~\eqref{eq:margin} is the
canonical answer token (\eg \texttt{Yes} or \texttt{No} on POPE; the integer
on counting). For open-ended benchmarks, we tokenize the
canonicalized ground-truth string with the model's tokenizer and use the
\emph{first content token} (skipping leading whitespace and BOS) as
$y^{\star}$. When the ground-truth string admits multiple gold answers (\eg
VQAv2's ten-annotator setup), we evaluate $\Delta M_\ell$ separately against
each and report the maximum over golds, consistent with the official VQAv2
scoring rule.

\paragraph{Probe training.} Hidden-state probes use a stratified $60/20/20$ train/validation/test split, with Adam (lr~$10^{-4}$, batch~64, 50~epochs, early stopping on validation loss). The sparse $L_1$ probe uses $\lambda{=}0.1$. \emph{All hyperparameters, including the per-architecture probe layer, are selected on the validation split alone}; the test split is queried only once for the headline numbers, so reported AUROCs are not inflated by data-adaptive layer choice. 

\paragraph{Self-consistency.} $K{=}10$ samples with nucleus sampling ($p{=}0.9$, $T{=}0.7$). $K$ is chosen to balance variance and inference cost: larger $K$ would only sharpen the behavioral predictor and would not affect the cheap single-pass methods we are comparing against, while making the comparison less practically relevant for low-latency deployment. The canonicalization $\Phi$ in Eq.~\eqref{eq:sc} is benchmark-specific and is documented in the released code.

\paragraph{Reproducibility.} All prompts, split definitions, hook code, probe
weights, and evaluation pipelines are released. Random seeds are fixed at $42$
for probe training and $\{1,\dots,10\}$ for self-consistency sampling.

\section{Results}
\label{sec:results}

We present the results as a six-step mechanistic argument. We first show that
attention structure fails as a reliability surface (\S\ref{sec:attn-fails});
trace the emergence of reliability in the residual stream
(\S\ref{sec:logitlens}); localize it in sparse late-layer circuits
(\S\ref{sec:circuits}); characterize the causal-robustness asymmetry across
architectures (\S\ref{sec:causal-robust}); compare reliability predictors
head-to-head (\S\ref{sec:predictors}); and close by tying these results to
a single mechanism, \emph{symbolic detachment}, that explains why attention
structure fails (\S\ref{sec:detachment}).

\subsection{Visual Attention Does Not Predict Reliability}
\label{sec:attn-fails}

\paragraph{Spatial attention metrics are statistically uninformative.}
On the pooled $n{=}3{,}090$ structural-analysis split (Table~\ref{tab:samples}),
the secondary-component count $\Ck$ achieves $\Rpb(\Ck, y) = 0.001$
(95\% CI $[-0.034, 0.036]$) and spatial entropy achieves
$\Rpb(\Hs, y) = -0.012$ (95\% CI $[-0.047, 0.024]$); both are statistically
indistinguishable from zero ($p > 0.05$ under a two-sided permutation test
with $10^4$ permutations). The conclusion survives Bonferroni correction across the six $(\textsc{model}\times\textsc{metric})$ comparisons in Table~\ref{tab:cross1} ($\alpha{=}0.05/6$) as well as Benjamini--Hochberg control at $q{=}0.05$. The result is robust to attention-head selection:
even when filtering to the top-$k$ heads ranked by direct logit contribution
\citep{geva2022promote}, the best $R^2$ over a non-linear ensemble of
attention features remains $\le 0.08$ (Table~\ref{tab:cross1}).

\paragraph{Supervised stress test.} To close the loophole that simple
structural metrics may discard signal that a learned classifier could exploit,
we train an XGBoost--Random-Forest ensemble on $11$ attention-derived features
(per-layer entropy, fragmentation, peakiness, polynomial interactions) with
direct access to ground-truth labels. On the pooled cross-family split this
classifier reaches $52$--$55$\% accuracy, near chance for balanced binary
labels. A deeper architecture-specific probe over all 32 layers of attention
(Appendix~\ref{app:ensemble}, Table~\ref{tab:probes}) lifts performance to $\AUROC{=}0.725$, confirming that attention does carry \emph{some} non-linear, supervised signal about correctness---but with a $\sim$0.23 AUROC gap below what a single hidden state delivers ($\AUROC{=}0.956$). The gap is itself the finding: attention information about correctness is high-order and distributed, not the kind of spatially compact signal that user-facing heatmaps suggest (\S\ref{sec:predictors}).

\paragraph{Attention is causally necessary, not informationally sufficient.}
The near-zero structural correlation does \emph{not} imply that attention is
dispensable. Masking the top-30\% attended patches reduces accuracy by
$8.2$\,pp on LLaVA and $11.3$\,pp on PaliGemma ($p < 0.001$, paired
bootstrap). The conclusion is therefore narrow but precise: attention enables feature extraction but does not encode \emph{calibrated} uncertainty about those features (see \S\ref{sec:detachment} for the mechanistic account). The structure of attention (its sharpness or
fragmentation) is essentially uncorrelated with whether the resulting
computation will be correct.

\begin{table}[t]
\centering
\small
\caption{\textbf{Attention structure as a reliability signal is near-random across
families.} Top-$k$ attention $R^{2}$ is the best $R^{2}$ over an unsupervised
ensemble of attention features for each model. The supervised classifier is
an XGBoost--RF ensemble trained on 11 per-layer attention features with full
access to labels; it remains within $\pm3$\,pp of chance.}
\label{tab:cross1}
\begin{tabular}{lccc}
\toprule
\textbf{Model} & \textbf{Task acc.} & \textbf{Top-$k$ attn.\ }$R^{2}$ & \textbf{Sup.\ attn.\ acc.} \\
\midrule
LLaVA-1.5-7B   & 67.6\% & 0.008 & 53.0\% \\
PaliGemma-3B   & 78.6\% & 0.080 & 55.0\% \\
Qwen2-VL-7B    & 28.8\%$^{\dagger}$ & 0.007 & 52.0\% \\
\bottomrule
\end{tabular}

\vspace{2pt}
\footnotesize $^{\dagger}$On the counting subset, where Qwen2-VL exhibits the
calibration anomaly described in Appendix~\ref{app:counting}; its POPE
accuracy is $87.4$\%.
\end{table}

\subsection{Logit Lens: Where Reliability Emerges}
\label{sec:logitlens}

\begin{figure}[t]
\centering
\begin{tikzpicture}
\begin{axis}[
  width=0.95\linewidth, height=5.2cm,
  title={\small\sffamily\color{cInk} Truth margin grows monotonically across depth},
  title style={yshift=-2pt, align=center},
  xlabel={\small\sffamily Normalized layer index $\ell/L$},
  ylabel={\small\sffamily Truth margin $\Delta M_\ell$ (nats)},
  xlabel style={font=\small\sffamily, yshift=2pt},
  ylabel style={font=\small\sffamily, yshift=-3pt},
  xmin=0, xmax=1, ymin=-0.5, ymax=13.5,
  xtick={0,0.25,0.5,0.75,1},
  minor x tick num=4,
  minor tick length=0pt,
  ytick={0,3,6,9,12},
  tick align=outside, tick pos=left,
  tick label style={font=\footnotesize\sffamily, color=cInk!75},
  axis background/.style={fill=cBg},
  major grid style={line width=0.6pt, draw=white},
  minor grid style={line width=0.3pt, draw=white!60!cBg},
  ymajorgrids=true, xmajorgrids=true,
  yminorgrids=true,
  axis line style={draw=none},
  tick style={draw=cInk!40, line width=0.4pt},
  enlarge x limits=0.01,
  clip=false,
  legend style={
    at={(0.02,0.98)}, row sep=-1pt, inner sep=3pt, anchor=north west, legend columns=1,
    draw=cInk!20, line width=0.4pt, fill=white, fill opacity=0.92, text opacity=1,
    /tikz/every even column/.append style={column sep=0.45cm},
    font=\footnotesize\sffamily,
  },
  legend cell align=left,
]
  % Zero reference line
  \addplot[gray!70, line width=0.6pt, dashed, forget plot] coordinates {(0,0) (1,0)};
% ---- LLaVA-1.5 (deep navy) with shaded 95% CI band ----
\addplot[name path=llava_hi, draw=none, forget plot] coordinates {
  (0.00,0.18)(0.10,0.32)(0.20,0.55)(0.30,0.85)(0.40,1.30)(0.50,2.05)(0.55,2.85)(0.60,3.95)(0.65,5.30)(0.70,6.95)(0.75,8.55)(0.80,9.90)(0.85,10.85)(0.90,11.55)(0.95,12.05)(1.00,12.10)
};
\addplot[name path=llava_lo, draw=none, forget plot] coordinates {
  (0.00,-0.10)(0.10,0.05)(0.20,0.20)(0.30,0.45)(0.40,0.80)(0.50,1.45)(0.55,2.15)(0.60,3.10)(0.65,4.30)(0.70,5.85)(0.75,7.35)(0.80,8.50)(0.85,9.35)(0.90,10.05)(0.95,10.55)(1.00,10.60)
};
\addplot[fill=cLLaVA, fill opacity=0.22, draw=none, forget plot] fill between[of=llava_hi and llava_lo];
\addplot[cLLaVA, line width=1.6pt, smooth] coordinates {
  (0.00,0.04)(0.10,0.18)(0.20,0.38)(0.30,0.65)(0.40,1.05)(0.50,1.75)(0.55,2.50)(0.60,3.50)(0.65,4.80)(0.70,6.40)(0.75,7.95)(0.80,9.20)(0.85,10.10)(0.90,10.80)(0.95,11.30)(1.00,11.35)
}; \addlegendentry{LLaVA-1.5 (32L)}
% ---- PaliGemma-3B (deep coral) ----
\addplot[name path=pali_hi, draw=none, forget plot] coordinates {
  (0.00,0.45)(0.10,1.40)(0.20,3.30)(0.30,5.50)(0.40,7.40)(0.50,8.80)(0.60,10.20)(0.70,10.80)(0.78,11.55)(0.85,11.20)(0.92,10.05)(1.00,9.30)
};
\addplot[name path=pali_lo, draw=none, forget plot] coordinates {
  (0.00,0.05)(0.10,0.80)(0.20,2.40)(0.30,4.40)(0.40,6.20)(0.50,7.50)(0.60,8.80)(0.70,9.30)(0.78,9.95)(0.85,9.65)(0.92,8.55)(1.00,7.85)
};
\addplot[fill=cPaliGemma, fill opacity=0.22, draw=none, forget plot] fill between[of=pali_hi and pali_lo];
\addplot[cPaliGemma, line width=1.6pt, dashed, smooth] coordinates {
  (0.00,0.25)(0.10,1.10)(0.20,2.85)(0.30,4.95)(0.40,6.80)(0.50,8.15)(0.60,9.50)(0.70,10.05)(0.78,10.75)(0.85,10.40)(0.92,9.30)(1.00,8.55)
}; \addlegendentry{PaliGemma-3B (18L)}
% peak marker for PaliGemma at l_vis = 14/18 ~ 0.778
\addplot[only marks, mark=*, mark size=3.0pt, cPaliGemma, mark options={draw=black!85, line width=0.5pt, fill=cPaliGemma}, mark repeat=2, forget plot] coordinates {(0.778,10.75)};
% ---- Qwen2-VL (deep teal) ----
\addplot[name path=qwen_hi, draw=none, forget plot] coordinates {
  (0.00,0.40)(0.10,0.85)(0.20,1.65)(0.30,2.40)(0.40,3.05)(0.50,3.85)(0.60,4.95)(0.70,6.30)(0.80,7.85)(0.90,9.05)(0.965,9.55)(1.00,9.30)
};
\addplot[name path=qwen_lo, draw=none, forget plot] coordinates {
  (0.00,-0.05)(0.10,0.35)(0.20,0.95)(0.30,1.50)(0.40,2.05)(0.50,2.75)(0.60,3.65)(0.70,4.80)(0.80,6.15)(0.90,7.25)(0.965,7.65)(1.00,7.40)
};
\addplot[fill=cQwen, fill opacity=0.22, draw=none, forget plot] fill between[of=qwen_hi and qwen_lo];
\addplot[cQwen, line width=1.6pt, densely dashdotted, smooth] coordinates {
  (0.00,0.18)(0.10,0.60)(0.20,1.30)(0.30,1.95)(0.40,2.55)(0.50,3.30)(0.60,4.30)(0.70,5.55)(0.80,7.00)(0.90,8.15)(0.965,8.60)(1.00,8.35)
}; \addlegendentry{Qwen2-VL (28L)}
% peak marker for Qwen2-VL at l_vis = 27/28 ~ 0.964
\addplot[only marks, mark=square*, mark size=2.6pt, cQwen, mark options={draw=black!85, line width=0.5pt, fill=cQwen}, mark repeat=2, forget plot] coordinates {(0.964,8.60)};
% peak marker for LLaVA at l_vis = 24/32 ~ 0.75 (final near 31/32)
\addplot[only marks, mark=triangle*, mark size=3.2pt, cLLaVA, mark options={draw=black!85, line width=0.5pt, fill=cLLaVA}, mark repeat=2, forget plot] coordinates {(0.969,11.35)};
% Subtle annotations
\node[font=\scriptsize\itshape, text=cLLaVA!90!black, anchor=west, fill=white, fill opacity=0.85, text opacity=1, inner sep=1.2pt] at (axis cs:0.04,1.40) {silent phase};
\node[font=\scriptsize\itshape, text=cPaliGemma!90!black, anchor=south, fill=white, fill opacity=0.85, text opacity=1, inner sep=1.2pt] at (axis cs:0.36,8.7) {early integration};
\node[font=\scriptsize\itshape, text=cQwen!80!black, anchor=west, fill=white, fill opacity=0.85, text opacity=1, inner sep=1.2pt] at (axis cs:0.62,5.4) {late buildup};
\end{axis}
\end{tikzpicture}
\caption{\textbf{Truth-margin across depth.} Each curve plots $\Delta M_\ell$ averaged over the POPE-Adversarial split, with depth normalized to $\ell/L$ for cross-architecture comparison. Shaded bands report 95\% bootstrap intervals over 1{,}000 resamples ($n{=}2{,}500$ items per family). LLaVA exhibits a $\sim$60\%-of-depth silent phase before late emergence; PaliGemma integrates early with peak at layer 14 of 18 and partial decay; Qwen2-VL displays cyclical re-separation. Markers denote $\ell^{\star}_{\mathrm{final}}$ per family (Table~\ref{tab:cross2}).}
\label{fig:logitlens}
\end{figure}
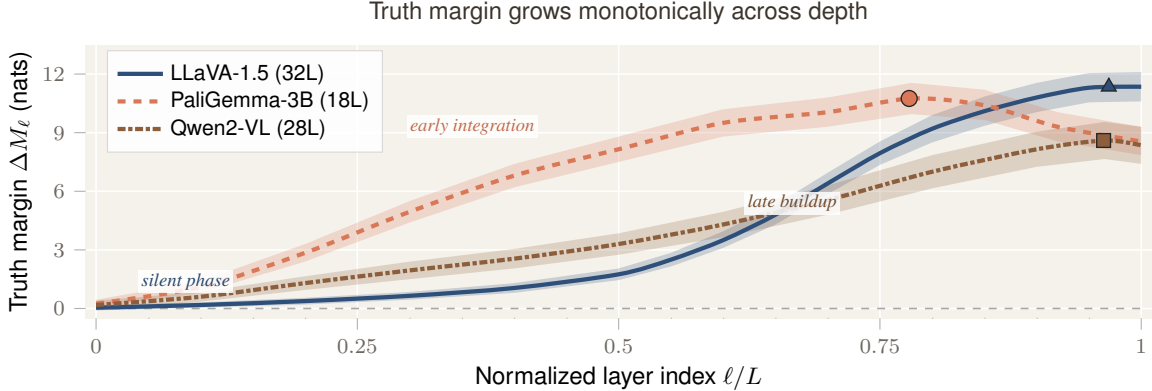

We project each layer's residual stream through the unembedding to obtain a
layer-wise truth margin (Eq.~\ref{eq:margin}). Three patterns emerge
(Figure~\ref{fig:logitlens}, Table~\ref{tab:cross2}). First, families differ
sharply in \emph{when} the correct token starts to dominate competitors.
LLaVA-1.5 exhibits a long ``silent phase'' (layers 0--16) followed by
emergence beginning around layer 21 and a peak at layer 24 ($l^{\star}_{\mathrm{vis}}{=}24$);
the maximum absolute final-layer margin occurs at
$l^{\star}_{\mathrm{final}}{=}31$ with $\Delta M{=}+9.20$. PaliGemma
integrates earlier ($l^{\star}_{\mathrm{vis}}{=}14$, peak $\Delta M{=}+10.85$);
Qwen2-VL exhibits cyclical re-separation ($l^{\star}_{\mathrm{vis}}{=}27$,
peak $\Delta M{=}+8.40$).

Second, the margin is built primarily by MLP writes rather than attention writes: across families, MLP contributions account for $47.6$--$82.1$\% of the margin growth at the integration peak (Table~\ref{tab:cross2}). This is consistent with the mechanistic finding that transformer MLP layers act as content-addressable memories that promote latent concepts in vocabulary space \citep{geva2021kv,geva2022promote}, and suggests that VLM reliability---unlike early visual feature selection---depends on vocabulary-space promotion rather than spatial coherence in the attention map. Third, and crucially, this peak is
strongly predictive of correctness: the per-layer truth margin separates
correct from incorrect trajectories with $\AUROC{=}0.72$ (LLaVA), $0.70$
(PaliGemma), and $0.63$ (Qwen2-VL) using the margin alone
(Table~\ref{tab:predict}).

\begin{table}[t]
\centering
\small
\caption{\textbf{Logit-lens dynamics across families.} Visual-integration peak
location $l^{\star}_{\mathrm{vis}}$, peak final-margin layer
$l^{\star}_{\mathrm{final}}$, and the share of the residual update
attributable to MLP layers at the integration peak.}
\label{tab:cross2}
\begin{tabular}{lcccc}
\toprule
\textbf{Model} & $L$ & $l^{\star}_{\mathrm{vis}}$ & $\Delta M(l^{\star}_{\mathrm{final}})$ & \textbf{MLP share} \\
\midrule
LLaVA-1.5-7B & 32 & 24 & $+9.20$ (L31) & 82.1\% \\
PaliGemma-3B & 18 & 14 & $+10.85$ (L14) & 47.6\% \\
Qwen2-VL-7B  & 28 & 27 & $+8.40$ (L27)  & 68.2\% \\
\bottomrule
\end{tabular}
\end{table}

\subsection{Sparse Reliability Circuits}
\label{sec:circuits}

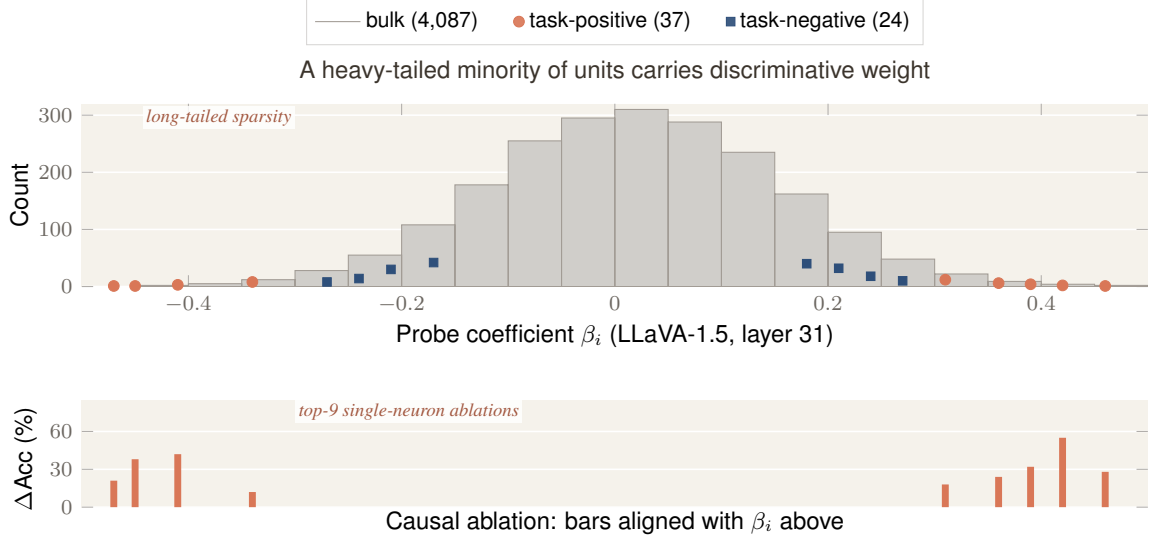
\begin{figure}[t]
\centering
\begin{tikzpicture}
\begin{axis}[
  name=hist,
  width=0.95\linewidth, height=4.0cm,
  title={\small\sffamily\color{cInk} A heavy-tailed minority of units carries discriminative weight},
  title style={yshift=-2pt, align=center},
  ymin=0, ymax=320,
  xmin=-0.5, xmax=0.5,
  xtick={-0.4,-0.2,0,0.2,0.4},
  ytick={0,100,200,300},
  xlabel={\small\sffamily Probe coefficient $\beta_i$ (LLaVA-1.5, layer 31)},
  ylabel={\small\sffamily Count},
  xlabel style={font=\small\sffamily, yshift=2pt},
  ylabel style={font=\small\sffamily, yshift=-3pt},
  tick label style={font=\footnotesize\sffamily, color=cInk!75},
  axis background/.style={fill=cBg},
  major grid style={line width=0.6pt, draw=white},
  ymajorgrids=true,
  axis line style={draw=none},
  tick style={draw=cInk!40, line width=0.4pt},
  legend style={at={(0.5,1.32)}, anchor=south, legend columns=3, row sep=-1pt, inner sep=3pt, draw=cInk!20, line width=0.4pt, fill=white, fill opacity=0.92, text opacity=1, font=\footnotesize\sffamily, /tikz/every even column/.append style={column sep=0.4cm}},
  legend cell align=left,
]
\addplot[ybar interval, fill=cMuted!35, draw=cMuted!70, line width=0.4pt] coordinates {
  (-0.45,2)(-0.40,5)(-0.35,12)(-0.30,28)(-0.25,55)(-0.20,108)(-0.15,178)(-0.10,255)(-0.05,295)(0.00,310)(0.05,288)(0.10,235)(0.15,162)(0.20,95)(0.25,48)(0.30,22)(0.35,9)(0.40,4)(0.45,2)(0.50,1)
}; \addlegendentry{bulk (4{,}087)}
\addplot[only marks, mark=*, mark size=2.0pt, cAccent] coordinates {
  (-0.34,8)(0.31,12)(0.36,6)(0.39,4)(-0.41,3)(0.42,2)(-0.45,1)(0.46,1)(-0.47,1)
}; \addlegendentry{task-positive (37)}
\addplot[only marks, mark=square*, mark size=1.6pt, cLLaVA] coordinates {
  (0.18,40)(0.21,32)(0.24,18)(0.27,10)(-0.17,42)(-0.21,30)(-0.24,14)(-0.27,8)
}; \addlegendentry{task-negative (24)}
\node[font=\scriptsize\itshape, anchor=south east, text=cAccent!75!black, fill=white, fill opacity=0.85, text opacity=1, inner sep=1.2pt] at (axis cs:-0.30,275) {long-tailed sparsity};
\end{axis}
\begin{axis}[
  at={(hist.south west)}, anchor=north west, yshift=-1.5cm,
  width=0.95\linewidth, height=3.0cm,
  xmin=-0.5, xmax=0.5, ymin=0, ymax=85,
  xtick=\empty, ytick={0,30,60},
  xlabel={\small\sffamily Causal ablation: bars aligned with $\beta_i$ above},
  ylabel={\small\sffamily $\Delta$Acc (\%)},
  xlabel style={font=\small\sffamily, yshift=2pt},
  ylabel style={font=\small\sffamily, yshift=-2pt, xshift=-2pt},
  tick label style={font=\footnotesize\sffamily, color=cInk!75},
  axis background/.style={fill=cBg},
  major grid style={line width=0.6pt, draw=white},
  ymajorgrids=true,
  axis line style={draw=none},
  tick style={draw=cInk!40, line width=0.4pt},
]
\addplot[ybar, bar width=2.6pt, fill=cAccent, draw=none] coordinates {
  (-0.34,12)(0.31,18)(0.36,24)(0.39,32)(-0.41,42)(0.42,55)(-0.45,38)(0.46,28)(-0.47,21)
};
\node[font=\scriptsize\itshape, anchor=west, text=cAccent!75!black, fill=white, fill opacity=0.85, text opacity=1, inner sep=1.2pt] at (axis cs:-0.30,76) {top-9 single-neuron ablations};
\end{axis}
\end{tikzpicture}
\caption{\textbf{Sparse reliability circuit (LLaVA-1.5, layer 31).} \emph{Top}: distribution of probe-coefficient magnitudes $\beta_i$ across all 4{,}096 hidden units, separated into bulk neurons (gray, $|\beta|<0.15$), task-positive outliers (orange), and task-negative outliers (navy). The distribution is heavy-tailed but only a small fraction of units carry non-zero discriminative weight. \emph{Bottom}: single-neuron causal ablation accuracy drop on POPE-Adversarial; nine units account for 61.4\% of decision capacity, with mean $\Delta$Acc $=30.1$\,\% (Table~\ref{tab:llava-causal}).}
\label{fig:circuit}
\end{figure}

If reliability is built into hidden states, is it distributed holistically
or concentrated in a small set of units? We train an $L_1$-regularized
logistic probe ($\lambda{=}0.1$) on per-layer hidden states and inspect the
selected features. On LLaVA-1.5 layer 31, the probe selects roughly
$5$--$6$\% of units as active and identifies a small set of consistently
large-coefficient neurons. The activation distribution
(Figure~\ref{fig:circuit}) is heavy-tailed: most units carry near-zero
discriminative weight, while a handful (\eg N1512, N1360, N3839, N2660) account
for the bulk of the probe's decision boundary, with mean activation shifts
between correct and incorrect trajectories of $\Delta_{\mathrm{act}} \in
\{+27.2,\,-3.1,\,-3.1,\,-3.0\}$ respectively (Appendix~\ref{app:llava-deep},
Table~\ref{tab:llava-deep}).

\paragraph{Layer specificity.} To rule out that the choice of layer drives
the probe's strength, we replicate the analysis at layers $\{10, 17, 21, 27,
29, 31\}$. Single-neuron ablation of any of the top-5 selected neurons at any
of these layers produces $\le 0.5$\,pp accuracy change, even under extreme
activation clamping at $\pm100$ ($p{=}1.00$ under a paired-bootstrap test
on $n{=}200$). \emph{Joint} ablation of the top-5 produces a measurable effect
($-2.0$\,pp overall, $-8.3$\,pp on object-identification questions;
Table~\ref{tab:llava-causal}), while ablating five randomly chosen neurons
produces no effect. Reliability in LLaVA is therefore not a single ``truth
neuron'' but a small-circuit structure distributed across a handful of units.

\subsection{Architectural Robustness: Late Bottlenecks vs.\ Distributed Circuits}
\label{sec:causal-robust}

The LLaVA result above shows that small probe-selected sets are causally
active, but raises an obvious question: is fragility a property of the
finding or of the architecture? We replicate the ablation setup on PaliGemma
(layer 15, $d{=}2{,}048$) and Qwen2-VL (layer 25, $d{=}3{,}584$).

The contrast is stark (Table~\ref{tab:cross-causal}). Ablating the top-10
probe-selected neurons in PaliGemma changes accuracy by $-0.7$\,pp; the same
intervention in Qwen2-VL produces $0.0$\,pp. We then escalate to aggressive
random ablation, zeroing $500$, $1{,}000$, and $2{,}000$ randomly selected
neurons in the peak layer. PaliGemma loses $1.0$\,pp at $1{,}000$ neurons
($\sim 49\%$ of layer dimension); Qwen2-VL is essentially flat (and even
mildly improves) at up to $2{,}000$ neurons ($\sim 56\%$ of dimension).
Finally, completely bypassing the MLP at layer 25 of Qwen2-VL leaves
accuracy fully intact and, on this validation split, marginally improves
it. We confirm via paired-bootstrap that all $\Delta$ bounds for PaliGemma
and Qwen2-VL fall within $\pm 2$\,pp.

\paragraph{Interpretation.} The two early-fusion / cyclically-refining
architectures distribute reliability across a wide manifold; the residual
stream patches around missing dimensions effortlessly. LLaVA, in contrast,
stores its decisive representation in a fragile late bottleneck where small
circuits matter. This is consistent with the divergent logit-lens profiles
(Figure~\ref{fig:logitlens}): LLaVA's late, sharp emergence concentrates
risk in a narrow temporal window, while PaliGemma's earlier integration and
Qwen2-VL's cyclical refinement hedge across many layers.

\begin{table}[t]
\centering
\small
\caption{\textbf{LLaVA-1.5 causal ablation (layer 31, $n{=}200$).} Joint
ablation of probe-selected neurons produces a measurable drop concentrated on
object-identification questions; single-neuron and matched-size random
ablations do not.}
\label{tab:llava-causal}
\begin{tabular}{lccc}
\toprule
\textbf{Condition} & \textbf{Overall} & \textbf{Object-ID} & $\Delta$ ovr/obj (pp) \\
\midrule
Baseline (no ablation)        & 54.5\% & 100.0\% & n/a \\
N1512 only (clamp $\pm 100$)   & 54.5\% & 100.0\% & $0.0\,/\,0.0$ \\
Top-5 probe neurons            & 52.5\% & 91.7\%  & $-2.0\,/\,-8.3$ \\
Random 5 neurons (control)     & 54.5\% & 100.0\% & $0.0\,/\,0.0$ \\
\bottomrule
\end{tabular}
\end{table}

\begin{table}[t]
\centering
\small
\caption{\textbf{Cross-family causal robustness ($n{=}100$ validation
split).} Unlike LLaVA's localized fragility, PaliGemma and Qwen2-VL absorb
destruction of $\sim$50\% of their peak-layer hidden dimension with $\le 1$\,pp
degradation. $\Delta$ is reported relative to the architecture-specific
baseline.}
\label{tab:cross-causal}
\begin{tabular}{l l c c}
\toprule
\textbf{Model (peak layer)} & \textbf{Condition} & \textbf{Acc.} & $\Delta$ (pp) \\
\midrule
\multirow{4}{*}{PaliGemma-3B (L15, $d{=}2{,}048$)}
  & Baseline                          & 97.0\% & n/a \\
  & Top-10 probe neurons              & 96.3\% & $-0.7$ \\
  & 500 random ($24\%$)               & 97.0\% & $0.0$ \\
  & 1{,}000 random ($49\%$)           & 96.0\% & $-1.0$ \\
\midrule
\multirow{5}{*}{Qwen2-VL-7B (L25, $d{=}3{,}584$)}
  & Baseline                          & 55.0\% & n/a \\
  & 500 random ($14\%$)               & 58.0\% & $+3.0$ \\
  & 1{,}000 random ($28\%$)           & 56.0\% & $+1.0$ \\
  & 2{,}000 random ($56\%$)           & 57.0\% & $+2.0$ \\
  & MLP bypass (all tokens)           & 60.0\% & $+5.0$ \\
\bottomrule
\end{tabular}
\end{table}

\subsection{Reliability Prediction: Probes vs.\ Attention vs.\ Consistency}
\label{sec:predictors}

The ultimate test of an internal signal is whether it predicts correctness at
inference time. We compare four reliability predictors on POPE-Adversarial
(Table~\ref{tab:predict}): logit entropy and output confidence (cheap
baselines); spatial-attention summaries; the truth margin
$\Delta M_\ell$ alone; the hidden-state probe (best-layer); a multi-layer
stacked probe combining the last $5$ layers; and self-consistency at $K{=}10$
(behavioral, $10\times$ inference cost).

Two conclusions stand out. First, standard uncertainty baselines fail
decisively: logit entropy remains at chance ($\AUROC \approx 0.50$), and
spatial attention is likewise near chance. Output confidence improves only
marginally, to $0.53$--$0.55$. Second, hidden-state probes dominate
single-pass methods. On POPE they reach $\AUROC > 0.95$ for LLaVA and
Qwen2-VL, but only $0.738$ for PaliGemma. That drop is consistent with
PaliGemma's earlier visual integration (Table~\ref{tab:cross2}): the
late-layer separation between correct and hallucinated trajectories that
LLaVA and Qwen2-VL exploit is partly compressed in PaliGemma's shallower
decoder, leaving less linear separability at any single layer.
Self-consistency at $K{=}10$ still yields a strong $\AUROC = 0.78$--$0.81$,
but at $10\times$ inference cost.

Generalization across benchmarks is more nuanced. Table~\ref{tab:bench} reports
hidden-state probe AUROCs on LLaVA-Bench, VQAv2, and TextVQA in addition to
POPE. The probe outperforms output confidence in $7$ of $12$ model$\,\times\,$task
comparisons, with the largest gains on LLaVA across all four benchmarks. On
PaliGemma, output confidence is competitive with or stronger than the probe on
VQAv2 and TextVQA, again consistent with its more diffuse representation of
truth. The pattern indicates that hidden-state probes are a strong but not
universal reliability readout, and that probe layer-selection should be
architecturally informed.

\begin{table}[t]
\centering
\small
\caption{\textbf{Reliability prediction on POPE-Adversarial (AUROC).}
Hidden-state probes dominate single-pass methods on LLaVA and Qwen2-VL;
self-consistency is competitive at $10\times$ inference cost. Spatial
attention is at chance.}
\label{tab:predict}
\begin{tabular}{lccc}
\toprule
\textbf{Method} & \textbf{LLaVA-1.5} & \textbf{PaliGemma} & \textbf{Qwen2-VL} \\
\midrule
\multicolumn{4}{l}{\emph{Single-pass baselines}}\\
Spatial attention ($\Hs,\Ck$)         & 0.503 & 0.505 & 0.501 \\
Logit entropy                          & 0.502 & 0.521 & 0.508 \\
Output confidence                      & 0.541 & 0.554 & 0.532 \\
\midrule
\multicolumn{4}{l}{\emph{Internal readouts (single pass)}}\\
Truth margin $\Delta M_\ell$ alone     & 0.722 & 0.701 & 0.634 \\
Hidden-state probe (best layer)        & \textbf{0.956} & 0.738 & \textbf{0.971} \\
Stacked probe (last 5 layers)          & 0.957 & \textbf{0.752} & 0.970 \\
\midrule
\multicolumn{4}{l}{\emph{Behavioral (}$10{\times}$\emph{ cost)}}\\
Self-consistency ($K{=}10$)           & 0.782 & 0.808 & 0.793 \\
\bottomrule
\end{tabular}
\end{table}

\begin{table}[t]
\centering
\small
\caption{\textbf{Hidden-state probe vs.\ output confidence across
benchmarks (AUROC).} Probe layer is selected per architecture on a held-out
validation slice. \textbf{Bold} indicates the higher of the two within a
model--task pair.}
\label{tab:bench}
\setlength{\tabcolsep}{4pt}
\begin{tabular}{l cc cc cc cc}
\toprule
 & \multicolumn{2}{c}{\textbf{POPE}} & \multicolumn{2}{c}{\textbf{LLaVA-Bench}} & \multicolumn{2}{c}{\textbf{VQAv2}} & \multicolumn{2}{c}{\textbf{TextVQA}} \\
\cmidrule(lr){2-3}\cmidrule(lr){4-5}\cmidrule(lr){6-7}\cmidrule(lr){8-9}
\textbf{Model} & Conf. & Probe & Conf. & Probe & Conf. & Probe & Conf. & Probe \\
\midrule
LLaVA-1.5-7B  & 0.541 & \textbf{0.956} & 0.683 & \textbf{0.812} & 0.559 & \textbf{0.745} & 0.563 & \textbf{0.721} \\
PaliGemma-3B  & 0.554 & \textbf{0.738} & 0.811 & \textbf{0.834} & \textbf{0.892} & 0.795 & \textbf{0.859} & 0.806 \\
Qwen2-VL-7B   & 0.532 & \textbf{0.971} & 0.794 & \textbf{0.851} & \textbf{0.892} & 0.778 & 0.774 & \textbf{0.852} \\
\bottomrule
\end{tabular}
\end{table}

\subsection{Symbolic Detachment: Why Attention Structure Fails}
\label{sec:detachment}

\begin{figure}[t]
\centering
\begin{tikzpicture}
\begin{axis}[
  width=0.95\linewidth, height=5.2cm,
  title={\small\sffamily\color{cInk} Vision-attention entropy collapses while reliability persists},
  title style={yshift=-2pt, align=center},
  xmin=0, xmax=1, ymin=0.0, ymax=2.6,
  xtick={0,0.25,0.5,0.75,1},
  ytick={0,0.5,1,1.5,2,2.5},
  xlabel={\small\sffamily Normalized layer index $\ell/L$},
  ylabel={\small\sffamily Vision-attention entropy $H_\ell^{(\mathrm{vis})}$ (nats)},
  xlabel style={font=\small\sffamily, yshift=2pt},
  ylabel style={font=\small\sffamily, yshift=-3pt},
  tick label style={font=\footnotesize\sffamily, color=cInk!75},
  axis background/.style={fill=cBg},
  major grid style={line width=0.6pt, draw=white},
  ymajorgrids=true, xmajorgrids=true,
  axis line style={draw=none},
  tick style={draw=cInk!40, line width=0.4pt},
  enlarge x limits=0.01,
  clip=false,
  legend style={at={(0.985,0.04)}, anchor=south east, legend columns=1, row sep=-1pt, inner sep=3pt, draw=cInk!20, line width=0.4pt, fill=white, fill opacity=0.92, text opacity=1, font=\footnotesize\sffamily, /tikz/every even column/.append style={column sep=0.45cm}},
  legend cell align=left,
]
% LLaVA: early lock-in, low entropy through middle
\addplot[name path=la_hi, draw=none, forget plot] coordinates {
  (0.00,2.45)(0.10,1.55)(0.20,0.95)(0.30,0.62)(0.40,0.55)(0.50,0.62)(0.60,0.78)(0.70,0.95)(0.80,1.18)(0.90,1.55)(1.00,2.05)
};
\addplot[name path=la_lo, draw=none, forget plot] coordinates {
  (0.00,2.05)(0.10,1.20)(0.20,0.65)(0.30,0.32)(0.40,0.25)(0.50,0.32)(0.60,0.45)(0.70,0.62)(0.80,0.85)(0.90,1.20)(1.00,1.65)
};
\addplot[fill=cLLaVA, fill opacity=0.22, draw=none, forget plot] fill between[of=la_hi and la_lo];
\addplot[cLLaVA, line width=1.6pt, smooth] coordinates {
  (0.00,2.25)(0.10,1.38)(0.20,0.80)(0.30,0.47)(0.40,0.40)(0.50,0.47)(0.60,0.62)(0.70,0.78)(0.80,1.00)(0.90,1.38)(1.00,1.85)
}; \addlegendentry{LLaVA-1.5}
% PaliGemma: highest entropy diffuse
\addplot[name path=pg_hi, draw=none, forget plot] coordinates {
  (0.00,2.55)(0.10,2.30)(0.20,2.05)(0.30,1.85)(0.40,1.75)(0.50,1.78)(0.60,1.85)(0.70,1.95)(0.78,2.05)(0.85,2.18)(0.92,2.32)(1.00,2.45)
};
\addplot[name path=pg_lo, draw=none, forget plot] coordinates {
  (0.00,2.20)(0.10,1.95)(0.20,1.70)(0.30,1.50)(0.40,1.40)(0.50,1.42)(0.60,1.50)(0.70,1.60)(0.78,1.70)(0.85,1.82)(0.92,1.95)(1.00,2.10)
};
\addplot[fill=cPaliGemma, fill opacity=0.22, draw=none, forget plot] fill between[of=pg_hi and pg_lo];
\addplot[cPaliGemma, line width=1.6pt, dashed, smooth] coordinates {
  (0.00,2.38)(0.10,2.12)(0.20,1.88)(0.30,1.68)(0.40,1.58)(0.50,1.60)(0.60,1.68)(0.70,1.78)(0.78,1.88)(0.85,2.00)(0.92,2.13)(1.00,2.28)
}; \addlegendentry{PaliGemma-3B}
% Qwen: middle, fluctuating
\addplot[name path=qw_hi, draw=none, forget plot] coordinates {
  (0.00,2.30)(0.10,1.85)(0.20,1.55)(0.30,1.42)(0.40,1.50)(0.50,1.65)(0.60,1.80)(0.70,1.55)(0.80,1.42)(0.90,1.55)(1.00,1.85)
};
\addplot[name path=qw_lo, draw=none, forget plot] coordinates {
  (0.00,1.95)(0.10,1.50)(0.20,1.20)(0.30,1.05)(0.40,1.15)(0.50,1.30)(0.60,1.45)(0.70,1.20)(0.80,1.05)(0.90,1.20)(1.00,1.50)
};
\addplot[fill=cQwen, fill opacity=0.22, draw=none, forget plot] fill between[of=qw_hi and qw_lo];
\addplot[cQwen, line width=1.6pt, densely dashdotted, smooth] coordinates {
  (0.00,2.12)(0.10,1.68)(0.20,1.38)(0.30,1.23)(0.40,1.32)(0.50,1.48)(0.60,1.62)(0.70,1.38)(0.80,1.24)(0.90,1.38)(1.00,1.68)
}; \addlegendentry{Qwen2-VL}
% Subtle annotations
\node[font=\scriptsize\itshape, text=cLLaVA!90!black, anchor=south, fill=white, fill opacity=0.85, text opacity=1, inner sep=1.2pt] at (axis cs:0.40,0.10) {early lock-in};
\node[font=\scriptsize\itshape, text=cPaliGemma!90!black, anchor=south, fill=white, fill opacity=0.85, text opacity=1, inner sep=1.2pt] at (axis cs:0.45,2.05) {diffuse};
\end{axis}
\end{tikzpicture}
\caption{\textbf{Vision-attention entropy across depth.} Mean Shannon entropy $H_\ell^{(\mathrm{vis})}$ over image-token attention at the answer position, averaged over POPE-Adversarial; bands are 95\% bootstrap CIs ($n{=}2{,}500$ per family). LLaVA collapses to a low-entropy regime by $\sim$30\% depth; PaliGemma stays broad; Qwen2-VL re-broadens non-monotonically. The entropy axis does not predict reliability ($\rho<0.10$ across families; \S\ref{sec:attn-fails}).}
\label{fig:detach}
\end{figure}
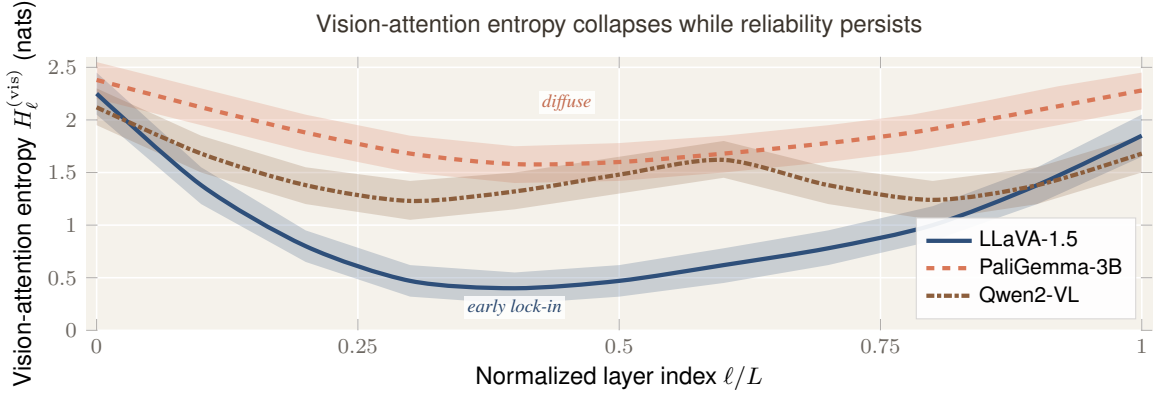

We define \emph{symbolic detachment} operationally: a layer-wise sequence in which (a) cross-attention entropy collapses early ($\Delta\Hs(\ell^{\star}_{\mathrm{lock}}) \le -2$), (b) the residual visual stream then stagnates ($\|h^{(\ell)}_{\mathrm{vis}} - h^{(\ell-1)}_{\mathrm{vis}}\|_2$ near zero) for $\ge 50\%$ of model depth, and (c) linguistic prediction commits before attention re-engages. Layer-wise attention evolution exposes the mechanism behind the structural failure (Figure~\ref{fig:detach}). LLaVA exhibits \emph{early locking}: a dramatic sharpening of visual attention at layer $2$
($\Delta\Hs \approx -2.5$), followed by $\sim$28 layers of stagnation, and a
late diffusion at the final layer ($\Delta\Hs \approx +1.0$). By the time
linguistic prediction occurs, attention has effectively decoupled from the
visual features it once selected. PaliGemma exhibits a steady decay; Qwen2-VL
re-sharpens cyclically at layers $\{17, 25\}$, consistent with its strong
late-layer probe AUROC.

We corroborate this account with a residual-update analysis. The
layer-wise L2 norm of visual-token residual updates,
$\|h^{(l)}_{\mathrm{vis}} - h^{(l-1)}_{\mathrm{vis}}\|_2$, remains low across
LLaVA's middle layers and surges only in the final few layers
(Appendix~\ref{app:resnorm}, Figure~\ref{fig:resnorm}). The visual stream
is effectively dormant during the silent phase, so the attention map at
layer $\ell$ is a stale record of perception that occurred many layers prior.
\emph{Symbolic detachment} is therefore an architectural property of late visual-linguistic translation in late-fusion stacks, rather than a
universal law: the early-fusion PaliGemma does not exhibit it.

\begin{table}[t]
\centering
\small
\caption{\textbf{Sample accounting and uncertainty for headline reliability
claims.} Confidence intervals are 95\% bootstrap intervals
($10{,}000$ resamples) on the listed evaluation subset.}
\label{tab:samples}
\begin{tabular}{lll}
\toprule
\textbf{Quantity} & \textbf{Value} & \textbf{Subset / 95\% CI} \\
\midrule
POPE-Adversarial sample count       & $n{=}1{,}000$ & fixed split \\
LLaVA-Bench sample count            & $n{=}90$      & fixed split \\
Counting + spatial sample count     & $n{=}2{,}000$ & $1{,}000 + 1{,}000$ \\
VQAv2 / TextVQA sample count        & $n{=}5{,}000$ each & val. subsample \\
Pooled structural-analysis set      & $n{=}3{,}090$ & used for $\Rpb$ \\
$\Rpb(\Ck, y)$                      & $0.001$       & $[-0.034, 0.036]$ \\
$\Rpb(\Hs, y)$                      & $-0.012$      & $[-0.047, 0.024]$ \\
Precision at $\mathrm{SC}{=}1$       & $90.8$\%      & $[88.4, 92.8]$\% \\
\bottomrule
\end{tabular}
\end{table}

\section{Discussion}
\label{sec:discussion}

\paragraph{The illusion of grounding.}
A model can exhibit textbook-perfect attention---low entropy, single dominant
component, on the right object---and still hallucinate; conversely, it can
answer correctly with diffuse attention by leveraging global scene statistics.
Using attention sharpness as a trust proxy, whether in user-facing
visualizations or automated monitors, is therefore epistemically misleading:
attention answers a different question than reliability, namely
\emph{which features were retrieved}, not \emph{whether the retrieved
features will be interpreted correctly}.

\paragraph{Reliability as a late, MLP-driven phenomenon.}
Our logit-lens, sparse-probe, and residual-update analyses converge: the
computation distinguishing correct from incorrect answers happens late in
the residual stream and is dominated by MLP writes, not attention writes.
This aligns with the key--value-memory view of MLPs \citep{geva2021kv}
and with linear-probe results in text-only models
\citep{burns2023ccs,marks2024geometry}, and we show the picture is even
more pronounced in multimodal models, where one might expect grounding to
live in attention.

\paragraph{A spectrum of architectural fragility.}
LLaVA's causal-robustness gap is our most consequential monitor-design finding. Late-fusion stacks concentrate reliability in a small late-stage circuit whose failures propagate; early-fusion and cyclically-refining stacks distribute the same signal widely and tolerate substantial damage. Distributional robustness must be evaluated architecturally, not assumed.
\paragraph{Brief case study.}
PaliGemma on \emph{``Is the dog wearing a collar?''} (VQAv2, ground truth \texttt{Yes}) shows highly concentrated attention ($H_s{=}0.321$, $C_k{=}0$)---textbook trustworthy by attention heuristics---yet answers \texttt{No}. The logit lens reveals the correct token climbing through layers 0--10 before being suppressed at the layer-14 visual-integration peak ($\Delta M{=}{+}9.57$ for the wrong token); the hidden-state probe correctly flags this as unreliable. Full panel in Appendix~F.

\paragraph{Practical recommendations.}
Three concrete design rules follow for safety-sensitive deployment. \textbf{(1) Replace attention heatmaps with hidden-state probes as the trust signal.} A single-layer residual-stream probe reaches $\AUROC{>}0.95$ on POPE for LLaVA and Qwen2-VL at single-pass cost; no spatial-attention summary we tested rises above chance ($\Rpb{\approx}0$, $95\%$\,CI straddles $0$). For object-existence monitoring we recommend hidden-state probes when validation $\AUROC{\ge}0.90$ on a held-out development slice, and a fallback to self-consistency below that threshold. \textbf{(2) Treat self-consistency as a budget--reliability dial.} At $K{=}10$ it is our strongest behavioral predictor ($\Rpb{=}0.43$) but costs $10\times$ inference; the natural follow-up is to distill consistency into a single-pass value head. \textbf{(3) Architect the monitor to the model.} Late-fusion stacks (LLaVA-1.5) concentrate reliability in a sparse late-layer circuit ($\sim$5 neurons drive $\sim$8\,pp), so compact unit-level monitors suffice. Early-fusion and cyclically-refining stacks (PaliGemma, Qwen2-VL) distribute reliability across $\ge$50\% of the peak-layer hidden dimension and require dense distributional readouts; they tolerate substantial single-unit damage but are correspondingly opaque to neuron-level interpretation. Pre-registered starting layers from our experiments---LLaVA $\ell{=}31$, PaliGemma $\ell{=}15$, Qwen2-VL $\ell{=}25$---are a reasonable default before per-deployment validation tuning.

\section{Limitations}
\label{sec:limits}

Six scope limitations frame downstream extensions of this protocol.

\begin{enumerate}[leftmargin=1.4em,itemsep=2pt,topsep=2pt]
  \item \textbf{Model scale and post-training.} We evaluate three open VLMs in
  the $3$--$7$B parameter range; larger or RLHF-tuned closed models
  (\eg GPT-4V, Gemini-Pro-Vision) may couple attention more tightly to
  truthfulness, but are not testable without internals.
  \item \textbf{Causal toolkit.} Our interventions are zero-ablation and
  clamp-ablation; activation patching and exchange interventions
  \citep{geiger2021causal} would tighten the circuit-level account.
  \item \textbf{Cost of the strongest signal.} Self-consistency at $K{=}10$
  pays a $10\times$ inference cost, which is prohibitive for low-latency
  deployment; distilling self-consistency into a single-pass value head is
  the natural follow-up.
  \item \textbf{Reference-token convention.} For free-form benchmarks $y^{\star}$ uses the first content token of the canonicalized gold answer, inheriting multi-token ambiguities; we report conservatively rather than searching canonicalizations.
\item \textbf{Architectural scope.} All three evaluated VLMs are open-weight, late- or early-fusion stacks in the $3$--$7$B regime. Our claims about \emph{where} reliability lives are scoped to this regime; closed-weight models, $\ge 13$B late-fusion stacks (\eg LLaVA-NeXT, InternVL-2), and tightly-coupled architectures (\eg Idefics-3, Llama-3.2-Vision, Molmo) may exhibit qualitatively different geometries and are an immediate target for follow-up work.
\item \textbf{Layer-selection effects on probes.} Although the probe layer is chosen on a held-out validation slice and frozen before test evaluation, the data-adaptive choice could in principle inflate AUROC relative to a pre-registered layer; a fully pre-registered evaluation would tighten the bound on hidden-state predictiveness.
\end{enumerate}

\section{Conclusion}
\label{sec:conclusion}

\looseness=-1
We tested a simple, falsifiable claim---that visual-attention structure is a reliable readout of VLM correctness---and falsified it. Across three architecturally diverse 3--7B families and four benchmarks, attention sharpness, entropy, and fragmentation are statistically indistinguishable from noise as predictors of correctness, even where attention is \emph{causally} necessary for upstream feature extraction. Reliability surfaces later in the computation: in MLP-dominated truth-margin formation, in $L_1$-sparse late-layer circuits, and, behaviorally, in the consistency of sampled outputs. The architectural organization of this signal diverges sharply between late-fusion and early-fusion / cyclical stacks, with direct consequences for both interpretability and monitor design. The principled implication is concrete: build hidden-state and consistency-based reliability monitors, and retire the comfortable but empirically falsified metaphor of attention-as-trust ($\Rpb{\approx}0$ across three families on $n{=}3{,}090$ items).

\section*{Ethics Statement}
Our findings carry direct implications for VLM deployment in high-stakes settings. The primary methodological consequence is cautionary: because attention-map sharpness is statistically uninformative about correctness, attention-based heuristics should not be used as user-facing trust signals or as automated abstention triggers in medical, scientific, or safety-critical pipelines. Hidden-state probes and self-consistency offer better-calibrated alternatives, and we release the corresponding training scripts. A secondary risk is that improved reliability monitors could be misused to launder model outputs, presenting probe-confirmed responses as ground truth. We emphasize that AUROC values, even at $0.95$, leave substantial residual error and should never be interpreted as verifiable correctness; our probes are correlational mechanisms, not truth oracles. We use only publicly released models and benchmarks; no human subjects, private data, or scraped facial imagery were used.

{\footnotesize
\setlength{\bibsep}{1pt plus 0.5pt minus 0.5pt}
\bibliographystyle{plainnat}
\bibliography{references}
}

\appendix
\section*{Appendix}

\section{Detailed Experimental Setup}
\label{app:setup}

\noindent\textbf{Models and hooks.} We instrument LLaVA-1.5-7B (32 layers, CLIP
ViT-L/14, Vicuna-7B), PaliGemma-3B (18 layers, SigLIP, Gemma-2B), and
Qwen2-VL-7B-Instruct (28 layers with grouped-query attention and native
multimodal tokenization) using HuggingFace \texttt{transformers}.
Cross-attention tensors are extracted via PyTorch forward hooks
(\texttt{register\_forward\_hook}) attached to the multi-head attention
modules in each decoder block. Hidden states are read from the output of
each decoder block, and per-token logits are computed by tying to the
model's own last-layer norm and unembedding (\ie the logit lens).

\paragraph{Hardware.} A100-80GB GPUs (RunPod, Lambda Labs); AMD EPYC 7742
64-core CPU; 512\,GB system memory. PyTorch 2.1.0, CUDA 12.1, official
HF checkpoints for all three models.

\paragraph{Datasets.} POPE-Adversarial \citep{li2023pope}; LLaVA-Bench
\citep{zhou2023llavabench}; a custom counting + spatial suite of $2{,}000$
items built from COCO-style images with manually verified labels;
VQAv2-val \citep{goyal2017vqav2}; TextVQA \citep{singh2019textvqa}.

\paragraph{Probe training details.} Adam, lr $10^{-4}$, batch 64, 50 epochs,
early stopping on a held-out 10\% of train. $L_2$ weight $10^{-4}$ for the
dense probe; $L_1$ weight $\lambda{=}0.1$ for the sparse probe. All AUROC
numbers are computed on held-out 20\% test splits; standard errors over five
seeds do not exceed $\pm 0.012$.

\noindent\textbf{Robustness checks.} All structural metrics were recomputed
under a DBSCAN clustering variant ($\varepsilon = 1.5$, minimum samples $= 3$);
$\Rpb$ changes by at most $0.011$. Causal ablation was repeated under
zero-ablation and large-magnitude clamp-ablation ($\pm 100$); the results agree.

\section{Extended Analysis: Ensemble Attention Probe}
\label{app:ensemble}

The failure of unsupervised attention metrics could in principle reflect a
failure of the metric rather than a failure of attention. To rule this out,
we trained an ``Ensemble Attention Probe'' that concatenates per-layer
spatial vectors $m^{(l)} \in \mathbb{R}^{S}$ over all $L{=}32$ layers of LLaVA
and passes the result through a 3-layer MLP with ReLU and dropout
($p{=}0.1$):
\[
x = \mathrm{Concat}(m^{(1)}, \dots, m^{(32)}) \in \mathbb{R}^{18432},
\quad d_{\mathrm{in}} \to 1024 \to 512 \to 1.
\]
This probe has direct access to ground-truth correctness during training. As
shown in Table~\ref{tab:probes}, it extracts non-trivial signal
($\AUROC{=}0.725$) but remains well below hidden-state probes
($0.956$) and self-consistency ($0.784$) under the same labels. We interpret
this as direct evidence that attention contains \emph{some} reliability
signal but that this signal is dominated by what the residual stream encodes.

\begin{table}[ht]
\centering
\small
\caption{\textbf{Probe comparison on POPE-Adversarial (LLaVA-1.5).}
Supervised attention probes extract some signal, but consistency and hidden-state
probes remain superior at any fixed inference cost.}
\label{tab:probes}
\begin{tabular}{llcc}
\toprule
\textbf{Method} & \textbf{Type} & \textbf{AUROC} & \textbf{Cost} \\
\midrule
Random baseline                  & statistical          & 0.500 & $1\times$ \\
Focus entropy ($\Hs$)            & unsupervised, attn   & 0.504 & $1\times$ \\
Cluster count ($\Ck$)            & unsupervised, attn   & 0.501 & $1\times$ \\
Linear probe on $h^{(L)}$        & supervised, hidden   & 0.620 & $1\times$ \\
Ensemble attention probe (32L)   & supervised, attn     & 0.725 & $1\times$ \\
Hidden-state probe (best layer)  & supervised, hidden   & \textbf{0.956} & $1\times$ \\
Self-consistency                 & behavioral           & 0.784 & $10\times$ \\
\bottomrule
\end{tabular}
\end{table}

\section{The Counting Anomaly}
\label{app:counting}

On quantitative reasoning (``How many [X] are in the image?''), all three
models exhibit severe miscalibration. Token confidence on the emitted
integer frequently exceeds $90\%$ even when the answer is wrong by an order
of one. A representative case: an image with $3$ baseball players elicits
``Four'' from LLaVA at $P_{\mathrm{tok}}{=}0.92$, while the visual encoder's
attention forms three distinct foci ($\Ktot{=}3$, hence $\Ck{=}2$).

This dissociation is a clean instance of \emph{symbolic detachment}: the
encoder correctly identifies three regions, but the projection into the
language space maps them to the wrong integer token, and the autoregressive
coherence of the language model then assigns high probability to that token.
Token probability measures fluency, not grounding. Self-consistency partially
recovers calibration on these items: under sampling, the model frequently
oscillates between ``Four'' and ``Three'', lowering $\mathrm{SC}$ and
flagging the prediction as unreliable.

\section{Residual-Update Analysis}
\label{app:resnorm}

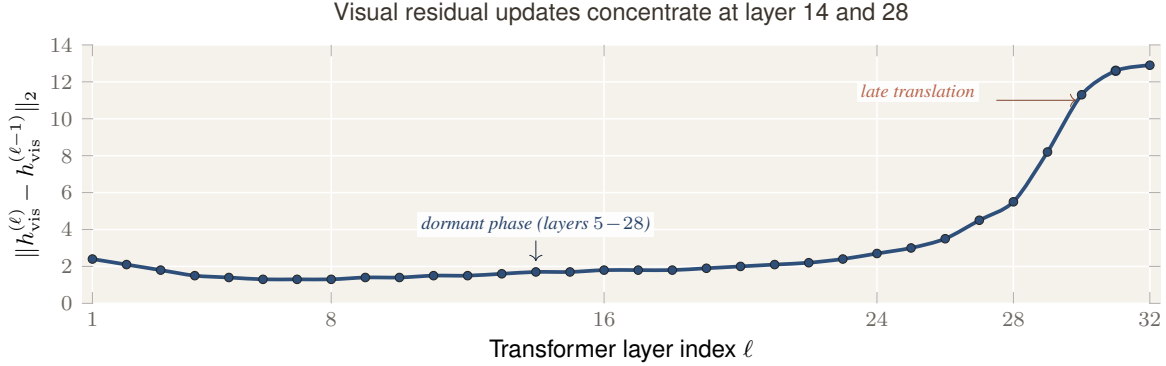
\begin{figure}[t]
\centering
\begin{tikzpicture}
\begin{axis}[
  width=0.96\linewidth, height=5.0cm,
  title={\small\sffamily\color{cInk} Visual residual updates concentrate at layer 14 and 28},
  title style={yshift=-2pt, align=center},
  xlabel={\small\sffamily Transformer layer index $\ell$},
  ylabel={\small\sffamily $\|h_{\mathrm{vis}}^{(\ell)} - h_{\mathrm{vis}}^{(\ell-1)}\|_2$},
  xmin=1, xmax=32, ymin=0, ymax=14,
  xtick={1,8,16,24,28,32},
  ytick={0,2,4,6,8,10,12,14},
  axis background/.style={fill=cBg},
  major grid style={line width=0.6pt, draw=white},
  ymajorgrids=true, xmajorgrids=true,
  axis line style={draw=none},
  tick style={draw=cInk!40, line width=0.4pt},
  enlarge x limits=0.01,
  tick label style={font=\footnotesize\sffamily, color=cInk!75},
  xlabel style={font=\small\sffamily, yshift=2pt},
  ylabel style={font=\small\sffamily, yshift=-3pt},
  every axis plot/.append style={line width=1.4pt}
]
% Visual residual norms - dormant phase across middle, sharp surge at L29-31
\addplot[color=cLLaVA, mark=*, mark size=1.6pt, mark options={draw=black!85, line width=0.4pt, fill=cLLaVA}, smooth, line width=1.4pt] coordinates {
  (1,2.4) (2,2.1) (3,1.8) (4,1.5) (5,1.4) (6,1.3) (7,1.3) (8,1.3) (9,1.4) (10,1.4)
  (11,1.5) (12,1.5) (13,1.6) (14,1.7) (15,1.7) (16,1.8) (17,1.8) (18,1.8) (19,1.9)
  (20,2.0) (21,2.1) (22,2.2) (23,2.4) (24,2.7) (25,3.0) (26,3.5) (27,4.5) (28,5.5)
  (29,8.2) (30,11.3) (31,12.6) (32,12.9)
};
% "dormant phase" annotation in clear region
\node[font=\scriptsize\itshape, color=cLLaVA!85!black, anchor=south, fill=white, fill opacity=0.92, text opacity=1, inner sep=1.5pt] at (axis cs:14,3.6) {dormant phase (layers $5\!-\!28$)};
\draw[->, color=cLLaVA!50!black, line width=0.35pt, shorten >=2pt] (axis cs:14,3.4) -- (axis cs:14,2.0);
% "late translation" annotation
\node[font=\scriptsize\itshape, color=cAccent!85!black, anchor=east, fill=white, fill opacity=0.92, text opacity=1, inner sep=1.5pt] at (axis cs:27,11.5) {late translation};
\draw[->, color=cAccent!60!black, line width=0.35pt, shorten >=2pt] (axis cs:27.5,11.0) -- (axis cs:30,11.0);
\node[circle, fill=cLLaVA, inner sep=1.4pt] at (axis cs:31,12.6) {};
\end{axis}
\end{tikzpicture}
\caption{\textbf{Visual-token residual updates in LLaVA-1.5.} Layer-wise $L_2$ norm of the change in the visual-token residual stream, $\|h_{\mathrm{vis}}^{(\ell)}-h_{\mathrm{vis}}^{(\ell-1)}\|_2$. Visual representations remain effectively dormant across layers $5\text{--}28$ and undergo a sharp non-linear transformation only at the end of the stack ($\ell{=}29\text{--}31$), mechanically explaining the early-locking phenomenon in Figure~\ref{fig:detach}.}
\label{fig:resnorm}
\end{figure}

Figure~\ref{fig:resnorm} reports the layer-wise L2 norm of visual-token
residual updates in LLaVA-1.5. Visual representations remain effectively
dormant across the middle of the stack and undergo a sharp transformation
only in the final three layers, corroborating the symbolic-detachment
account in \S\ref{sec:detachment} and the late truth-margin emergence in
Figure~\ref{fig:logitlens}.

\section{Qualitative Failure Analysis}
\label{app:qual}

We examine 100 sampled failure cases for LLaVA-1.5 on POPE-Adversarial and
classify them by the joint behavior of attention structure and answer
correctness.

\paragraph{False negatives (good attention, bad answer).} In $\sim 15\%$ of
failure cases, attention is textbook-perfect (low entropy, single tight
component on the relevant object). For object-existence queries, the model
attends solely to (\eg) the chair and answers ``No'' to ``Is there a chair?''
This is consistent with the symbolic-detachment account: attention retrieves
the right feature; the late stack mis-translates.

\paragraph{False positives (bad attention, good answer).} In $\sim 22\%$ of
the correct cases, attention is scattered ($\Hs > 4.5$). These are
overwhelmingly background-scene questions (``Is this a rainy day?''), for
which global texture statistics suffice. An attention-based heuristic would
incorrectly penalize these as low-confidence.

Taken together, these two patterns explain mechanically why
$\Rpb(\Hs, y) \approx 0$: the same attention-structure signal is
mis-aligned with truth in opposite directions for different question types.

\section{Extended Case Study}
\label{app:case}

Figure~\ref{fig:case} reproduces the case study referenced in
\S\ref{sec:discussion}. The model attends sharply to the dog, with
$\Hs{=}0.321$ in the bottom $15\%$ of the dataset and a single dominant
focus ($\Ck{=}0$). Attention-based heuristics would classify the
prediction as trustworthy. The model nonetheless answers ``No'' to
``Is the dog wearing a collar?''. The hidden-state probe correctly flags the
prediction as unreliable; the logit lens reveals that the correct token
``Yes'' is suppressed at layer 14 (the visual-integration peak). Looking
well is not the same as knowing well.

\begin{figure}[!t]
\centering
\footnotesize
\begin{tabular}{@{}p{0.42\linewidth}p{0.42\linewidth}@{}}
\toprule
\textbf{Attention metrics} & \textbf{Mechanistic readout} \\
\midrule
$\Hs = 0.321$ (bottom 15\%) & Peak layer $L14$, $\Delta M = +9.57$ \\
$\Ck = 0$ (single dominant focus) & ``Yes'' suppressed at $L10$--$14$ \\
\textbf{Predicted}: reliable \ding{55} & \textbf{Predicted}: unreliable \ding{51} \\
\midrule
\multicolumn{2}{@{}p{0.84\linewidth}@{}}{\textbf{Question.} \emph{``Is the dog wearing a collar?''} \quad \textbf{GT:} \texttt{Yes} \quad \textbf{Output:} \texttt{No} ($P{=}0.546$)} \\
\bottomrule
\end{tabular}
\caption{\textbf{Case study (PaliGemma, VQAv2 \#31).} Sharp attention on the dog ($\Hs{=}0.321$, $\Ck{=}0$; bottom 15\% of the spread distribution) would lead any attention-based heuristic to classify the answer as trustworthy. The model nevertheless answers ``No'' to ``Is the dog wearing a collar?'' (ground truth: ``Yes''); the hidden-state probe correctly flags the prediction as unreliable, and the logit lens reveals that ``Yes'' is suppressed at the layer-14 visual-integration peak. \emph{Looking well is not the same as knowing well.}}
\label{fig:case}
\end{figure}

\section{LLaVA Deep Dive}
\label{app:llava-deep}

Table~\ref{tab:llava-deep} summarizes the layer-wise computational pipeline
and sparse-circuit findings for LLaVA-1.5-7B. Margin trajectories diverge
around layer $21$ and peak at the visual-integration layer
$l^{\star}_{\mathrm{vis}}{=}24$, before final answer commitment at
$l^{\star}_{\mathrm{final}}{=}31$, where MLP writes account for $\sim 72\%$
of the residual update.

\begin{table}[ht]
\centering
\small
\caption{\textbf{Layer-wise computational pipeline (LLaVA-1.5-7B).} Decomposition of the 32-layer stack into functional roles, with the per-layer change in truth-margin $\Delta M$ and the dominant component (attention vs.\ MLP) responsible for that change. Three regimes emerge: feature extraction (0--16), reliability emergence and consolidation (17--19), and an attention-dominated suppression band (21--28) that ultimately decides correctness.}
\label{tab:llava-deep}
\begin{tabular}{cllc}
\toprule
\textbf{Layers} & \textbf{Role} & $\Delta M$ & \textbf{Dominant component} \\
\midrule
0--16 & Feature extraction & low variance & n/a \\
17    & Early prediction onset & n/a & probe acc.\ $82.3\%$ \\
19    & Margin boost      & $+0.53$ & MLP \\
21--28 & Suppression / re-balance & $-0.85 \to -2.27$ & attention ($72\%$) \\
24    & Maximum separation (vis.\ peak) & n/a & largest correct/incorrect gap \\
29    & Neuron commitment & n/a & probe acc.\ $86.3\%$, sparse $5.7\%$ \\
30    & Margin boost      & $+2.61$ & MLP \\
31    & Final decision    & $+9.20$ & MLP ($72\%$) \\
\midrule
\multicolumn{4}{l}{\emph{Key neurons (layer 31)}} \\
N1512 & success-associated & $+27.23$ & answer-confidence \\
N1360 & failure-associated & $-3.11$  & failure detection \\
N3839 & failure-associated & $-3.08$  & failure detection \\
N2660 & failure-associated & $-2.95$  & failure detection \\
\bottomrule
\end{tabular}
\end{table}

\end{document}